\documentclass[]{trbunofficial}
\usepackage{graphicx}

\usepackage[hidelinks]{hyperref}
\usepackage{amssymb}
\usepackage{bm}
\usepackage{amsmath}
\usepackage{multirow}
\usepackage{float}
\usepackage{color}

\usepackage{multirow}
\usepackage{multicol}
\usepackage{float}
\usepackage{subcaption}

\newcommand*\tabrule[1][]{%
   \if\relax\detokenize{#1}\relax
     \rule{\linewidth}{\arrayrulewidth}%
   \else
     \rule{#1}{\arrayrulewidth}%
   \fi
}

\AuthorHeaders{Zhao et al.}
\title{Distilling Black-Box Travel Mode Choice Model for Behavioral Interpretation}

\author{%
  \textbf{Xilei Zhao, Corresponding Author}\\
  Department of Civil and Coastal Engineering\\
  University of Florida\\
  Gainesville, FL 32603\\
  xilei.zhao@essie.ufl.edu\\
  \hfill\break
  \textbf{Zhengze Zhou}\\
  Department of Statistics and Data Science\\
  Cornell University\\
  301 Malott Hall, Ithaca, NY 14853\\
  zz433@cornell.edu\\
  \hfill\break
  \textbf{Xiang Yan}\\
  Taubman College of Architecture and Urban Planning\\
  University of Michigan\\
  2000 Bonisteel Blvd, Ann Arbor, MI 48109\\
  jacobyan@umich.edu\\
  \hfill\break
  \textbf{Pascal Van Hentenryck}\\
  H. Milton Stewart School of Industrial and Systems Engineering\\
  Georgia Institute of Technology\\
  755 Ferst Drive, NW, Atlanta, GA 30332\\
  pvh@isye.gatech.edu
}

\begin{document}
\maketitle

\section{Abstract}

Machine learning has proved to be very successful for making predictions in travel behavior modeling. However, most machine-learning models have complex model structures and offer little or no explanation as to how they arrive at these predictions. Interpretations about travel behavior models are essential for decision makers to understand travelers' preferences and plan policy interventions accordingly. Therefore, this paper proposes to apply and extend the model distillation approach, a model-agnostic machine-learning interpretation method, to explain how a black-box travel mode choice model makes predictions for the entire population and subpopulations of interest. Model distillation aims at compressing knowledge from a complex model (teacher) into an understandable and interpretable model (student). In particular, the paper integrates model distillation with market segmentation to generate more insights by accounting for heterogeneity. Furthermore, the paper provides a comprehensive comparison of student models with the benchmark model (decision tree) and the teacher model (gradient boosting trees) to quantify the fidelity and accuracy of the students' interpretations.

\hfill\break%
\noindent\textit{Keywords}: model distillation, travel mode choice, black-box, interpretation, heterogeneity
\newpage


\section{Introduction}

Applying machine learning to model travel behavior has become popular in the past several years. With much higher predictive accuracy compared to traditional random utility models, the machine-learning approach has become appealing for making predictions, especially at the individual level. However, many machine-learning models are black-box and are often deemed as uninterpretable; interpretation, on the other hand, is an important aspect of travel behavior modeling---to conduct policy analysis and facilitate decision making. 

Interpretable or explainable machine learning, an emerging field in Statistics and Computer Science, has attracted a lot of attention in the past two years. It aims at peeking into the black box to extract relevant knowledge and provides useful insights about the importance of various features and how they impact predictions. \citet{zhao2018modeling,zhao2019modeling} have explored applying several model-agnostic machine-learning interpretation methods, including variable importance, partial dependence plots \citep{friedman2001greedy}, individual conditional expectation plots \citep{goldstein2015peeking}, marginal effects, and elasticity to explain the black-box travel mode choice models. However, most of these methods are permutation-based, and they may produce misleading diagnostics, particularly when strong dependences exist among features \citep{hooker2019please}.

As an alternative, model distillation methods have been advocated by some researchers to offer more direct interpretations about black-box models \citep{zhou2018approximation, gibbons2013cad, hinton2015distilling}. 
Model distillation operates by developing intelligible \textit{student} models that mimic the predictions of the original black-box \textit{teacher} model. The model distillation approach may be able to provide direct explanations from the transparent student models, whose \textit{fidelity} (how well the mimic model's outputs agree with the teacher model's predictions) and \textit{accuracy} (how many test-set instances can be correctly classified) can be quantitatively measured. These insights then can be used to guide transportation planning and policy interventions. However, to the best of our knowledge, model distillation has not been applied to interpret black-box travel behavior models. This paper aims at addressing this research gap by using model distillation to interpret a black-box travel mode choice model that evaluates travelers' preferences for a new mobility-on-demand (MOD) transit system.

Moreover, model distillation applications have traditionally focused on interpreting predictions for  the entire population under evaluation (e.g., \cite{gibbons2013cad,papernot2016distillation}). However, in travel behavior modeling, researchers and practitioners may be more interested in specific segments of the population, such as car users, female population, and low-income population to name only a few \citep{yan2019mobility}. Therefore,  this paper proposes to integrate model distillation with market segmentation (a widely-adopted econometric method) to investigate the taste heterogeneity of the population and develop better-targeted policies for various groups under evaluation.

The unique contributions of this paper are summarized as follows:
\begin{itemize}
    \item Apply model distillation, an interpretable machine learning tool, to explain black-box travel mode choice models, aiming at drawing useful insights from complex machine-learning models for policy analysis and decision making;
    

    \item Integrate model distillation with market segmentation to generate more insights by accounting for heterogeneity and design better-targeted policies;
    
    \item Propose a methodological framework to guide machine-learning travel behavior modelers to implement the overall idea.
    
\end{itemize}

The rest of the paper is organized as follows. Section 2 reviews existing literature on travel mode choice modeling using traditional random utility models and machine learning. Section 3 describes the methodological framework, including the fundamentals of model distillation and the integration of model distillation with market segmentation. Section 4 applies the proposed framework to a case study of stated-preference data for a new MOD transit system in Ann Arbor, Michigan. Lastly, the paper concludes by summarizing the major findings and suggesting future research directions.

\section{Literature Review}

\subsection{Random Utility Models and Their Applications in Travel Behavior Modeling}

The random utility model is a class of econometric models based on random utility maximization theory \citep{ben1985discrete}, and has been widely applied to model travel mode choice since 1970s \citep{mcfadden1973conditional}. It assumes that each travel mode has its own utility, and an individual tends to choose the travel mode with the highest utility. 

There are various types of random utility models, and the simplest and most popular one is the multinomial logit (MNL) model. However, the MNL model is often challenged for its major assumption, i.e., the independence of irrelevant alternatives (IIA) property, and its inability to model taste heterogeneity among different individuals. In order to tackle these limitations, the nested logit model and mixed logit model were later developed \citep{ben1999discrete}. The nested logit model was specifically invented to account for the correlated alternatives (i.e., travel modes). The mixed logit model, as a more advanced random utlity model, does not rely on the IIA property and can accommodate individual heterogeneity \citep{hensher2003mixed}. 

Even though random utility models are based on sound behavioral assumptions and can offer intuitive interpretations, they often suffer from poor predictive capabilities, especially at the individual level \citep[e.g.,][]{xie2003work,omrani2015predicting,hagenauer2017comparative,zhao2018modeling}.

\subsection{Interpretable Machine Learning and Its Applications in Travel Behavior Modeling}

Machine learning, with flexible modeling structure and high predictive accuracy, has gained a lot of popularity in the research area of travel behavior modeling \citep[e.g.,][]{hagenauer2017comparative, lheritier2018airline, xie2003work, omrani2015predicting, wang2018machine}. For instance, \citet{xie2003work} used decision trees and artificial neural networks to model travel mode choice and showed improved predictive accuracy over the MNL model. Later, \citet{omrani2015predicting} compared the MNL model with the support vector machine, multi-layer perceptron network, and the radial basis function network, and found that the latter three machine-learning models have significant better predictive capability than the MNL model in terms of overall accuracy and accuracy for each travel mode. Then, \citet{hagenauer2017comparative} conducted a comprehensive comparison of multiple machine-learning algorithms for modeling travel mode choice and found that the random forest model outperformed all the other models in terms of prediction. 

However, these studies mainly focused on improving the predictive accuracy by using machine learning, with little discussion on how to interpret these machine-learning models for better decision-making. Fortunately, interpretable or explainable machine learning has drawn much attention in the fields of Statistics and Computer Science over the past two years \citep[e.g][]{murdoch2019interpretable,doshi2017towards,molnar,zhao2017causal,wager2018,athey2017beyond,zhou2018approximation}, in order to ``X-ray the black box'' and facilitate knowledge extraction from machine learning. In particular, \citet{zhao2018modeling} applied several popular machine-learning interpretation tools to model and explain travel mode choice for the entire population under evaluation. Subsequently, \citet{zhao2019modeling} looked into how to account for the taste heterogeneity in travel mode choice by applying existing and inventing new interpretable machine learning techniques. However, some of the tools used in these two studies are permutation-based, and as pointed out by \citet{hooker2019please}, they may generate misleading results, when there are strong dependencies among features.


In contrast, model distillation is another major approach for interpretable machine learning that can overcome the limitations of permutation-based methods and offer transparent explanations. For instance, \citet{bucilua2006model} trained compact artificial neural nets to mimic the functioned learned by an ensemble methods. They used the term model compression instead of distillation as their main purpose is to compress the large model without significant loss in prediction performance. A successful application was presented in \citep{gibbons2013cad}, where the authors resorted to decision trees as student model for shortening medical questionnaires, reducing data collection burden and improving interpretation. \citet{hinton2015distilling} further developed model distillation using a different compression
technique. Most recently, \citet{zhou2018approximation} examined statistical stability in model
distillation using regression trees. They developed tests to stabilize the choice of splits and a stopping rule to indicate how deep the tree should be built. Nevertheless, the model distillation approach has not been used to examine how black-box travel behavior models arrive at these predictions and how to draw useful insights for decision making.

\section{Methodology}

This section first introduces the fundamentals of model distillation. It then discusses how to integrate model distillation with market segmentation to account for heterogeneity across different subpopulations. Lastly, it proposes the overall methodological framework. 


\subsection{Fundamentals of Model Distillation}

It is difficult to explain how black-box machine-learning models work. The task of extracting knowledge from a black-box model is to summarize, in a comprehensible way, the knowledge learned by a black-box model during training \citep{adadi2018peeking}. A popular technique for knowledge extraction of black-box models is called \textit{model distillation}. Distillation can be seen as a model compression process to transfer knowledge from a large, complex \textit{teacher model} to an interpretable \textit{student model} \citep{hinton2015distilling, tan2018distill}. In other words, model distillation gains insights and draws conclusions about the teacher model by interpreting the student model. 

Model distillation is typically achieved by using the teacher model to make predictions for the unlabeled samples (either new unlabeled data or training data with labels discarded), and then training the student model to mimic the teacher's predictions \citep{tan2018distill}. Furthermore, in order to transfer the generalization ability of the teacher to a student model, one can use the soft predictions (i.e., the predicted class probabilities) of the teacher model as the target to train the student model. Compared with a hard label, the predicted class probabilities can offer much richer information about similarity between the classes and reveal teacher model's confidence in prediction \citep{hinton2015distilling}. 



In order to measure the student model's performance in approximating the teacher's model (referred to as the fidelity), different evaluation methods can be used. One commonly-adopted method is the R-squared measure (or $R^2$) \citep{molnar}:

$$R^2 = 1 - \frac{\sum_{i=1}^n (\hat{Y}_i-\widetilde{Y}_i)^2}{\sum_{i=1}^n (\hat{Y}_i -\bar{\hat{Y}})^2},$$
\\

\noindent where $\hat{Y}_i$ is the prediction for the $i$th observation of the teacher model, $\widetilde{Y}_i$ is the prediction for the $i$th observation of the student model, $\bar{\hat{Y}}$ is the mean of the teacher model's predictions, and $n$ is the total number of observations. $R^2$ quantifies how much variance can be explained by the student model. If $R^2$ is close to 1, it indicates that the student model can approximate the teacher model very well; in contrast, if it is close to 0, it shows poor performance of the student model to interpret the teacher model. 




The student model is trained to interpret the teacher model, so the interpretation gained from the student model is all about exploring the internal prediction mechanism of the black-box model. The main question then is whether this interpretation can be used and trusted for policy analysis and decision making? This paper argues that this depends on three major conditions. The first condition is the predictive capability of the teacher model. Without an accurate teacher model, the interpretation of the student model may become irrelevant.
The second condition is the fidelity of the student model. If a student model is highly faithful to the teacher,  the interpretations gained from the teacher can be trusted with high confidence.   
The third condition is the predictive accuracy of the student model. The student model is typically evaluated against in-sample $R^2$ for its fidelity \citep{molnar}; however, a student model may overfit the training data (showing high $R^2$) but lack out-of-sample predictive capability. This paper argues that it is necessary to leverage the ground-truth outcome information of a separate test set to evaluate the student model's out-of-sample predictive accuracy.

Within model distillation, any interpretable model can be used as student models, which offers lots of flexibility for further analysis and explanation. Common student model classes include (generalized) linear models \citep{tan2018distill}, generalized additive models \citep{lou2012intelligible} and decision trees \citep{gibbons2013cad,zhou2018approximation}. Decision trees are the focus in this paper: They present an intelligible graphical representation and may automatically capture complex high-dimensional data, making them appealing as student models. Moreover, their rule-based structure can be directly fed into some activity-based travel demand model, such as ALBATROSS \citep{arentze2004learning} and AMOS \citep{pendyala1998application}.

\subsection{Integrating Model Distillation with Market Segmentation}

In travel mode choice modeling, understanding taste heterogeneity is an important research topic within the random utility framework. From a modeler's standpoint, taste heterogeneity can be divided into two parts, including observed and unobserved heterogeneity. The observed heterogeneity mainly results from the observed individual features and/or their interactions with the level-of-service features, while the unobserved heterogeneity is usually caused by the unobserved individual features, such as the individuals' intrinsic bias towards different travel modes and/or their different degrees of sensitivity to the level-of-service features \citep{bhat2000incorporating}. In particular, the observed heterogeneity can be modeled by adding observed individual socio-demographic or behavioral features as alternative specific variables and/or by modeling interactions between level-of-service features and observed individual features, e.g., by applying a market segmentation approach or introducing interaction terms \citep{bhat2000incorporating}.

In the market segmentation approach, sub-populations from a market are selected according to the observed feature(s) and thus declared as ``segments.'' Market segmentation aims at analyzing a manageable number of groups that share well-defined underlying features and generating more innovative and better-targeted strategies for different groups \citep{anable2005complacent}. \citet{zhao2019modeling} have demonstrated that applying the market segmentation approach to partial dependence plots and individual conditional expectation plots can help extract additional insights for various market segments.

Therefore,  this paper proposes to apply the market segmentation approach to model distillation, aiming at better understanding heterogeneous population and generating more effective policies for the subpopulations of interest. 



\subsection{Proposed Methodological Framework}

\begin{figure}[!t]
    \centering
    \includegraphics[width=16cm]{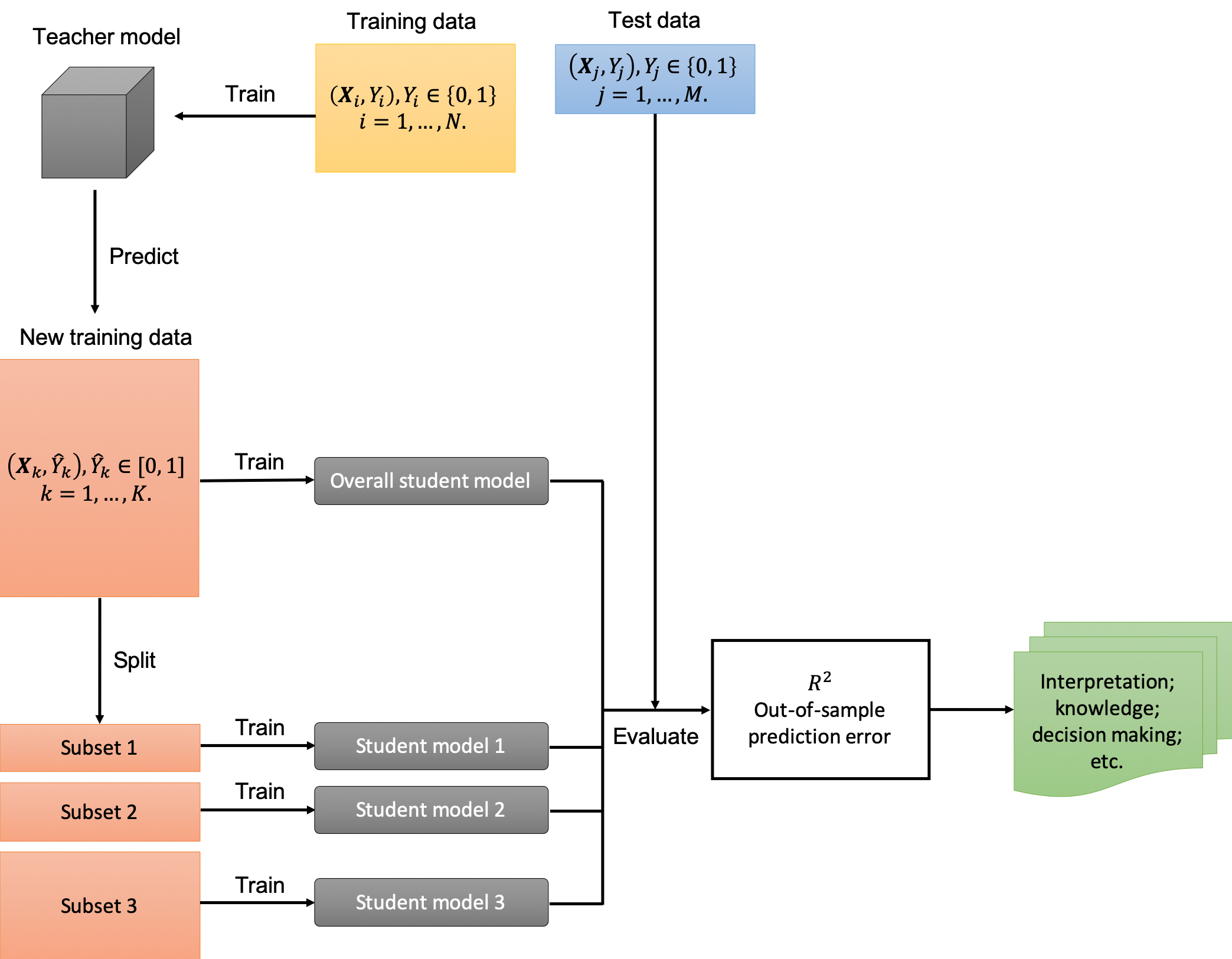}
    \caption{Illustration of the proposed methodological framework.}
    \label{fig:framework}
\end{figure}

This paper proposes an overall methodological framework to guide transportation planners and engineers to extract insights from black-box travel behavior models, which is illustrated in Figure \ref{fig:framework}. 

It assumes the availability of a dataset that includes individual travel mode choice before and after a new mobility service is introduced. The $t^{th}$ observation of the dataset is represented as $(\boldsymbol{X}_t, Y_t), t = 1, 2, ..., T$, where $\boldsymbol{X}_t = [X_{t1}, ..., X_{tp}]$, a vector of $p$ features for individual $t$, $Y_t$ is the response variable, and $T$ is the total number of observations. To be specific, $\boldsymbol{X}_t$ include important information of each individual such as her socio-demographic information, her travel preference, and her level-of-service variables for each travel mode under evaluation. The response variable $Y_t$ is binary with 0 indicating that individual $t$ stays with her current mode and 1 if she switches to the new mobility option; that is to say, $Y_t \in \{0, 1\}$.

As shown in Figure \ref{fig:framework}, the original dataset is split into two separate subsets, with $N$ observations in the training set and $M$ observations in the test set, where $N + M = T$. The training set is used to train the black-box teacher model. Then, additional synthetic data are generated to form a new training dataset. The main reason for doing this step is because the decision trees (the focus of this paper) may be sensitive to small perturbations and lead to unstable model structures. Hence, as discussed in \citep{gibbons2013cad,zhou2018approximation}, synthetic data can be generated from a kernel density estimate based on the observed features. For the original training data along with the generated synthetic data, i.e., $\boldsymbol{X}_k, k = 1, ..., K$, the teacher model is used to make soft predictions (i.e., switching probabilities) for each instance, where the soft predictions can be denoted as $\hat{Y}_i \in [0,1]$. Thus, $\boldsymbol{X}_k$ and $\hat{Y}_k$ form the new training set, $D = \{\boldsymbol{X}_k, \hat{Y}_k\}_{k = 1}^{K}$, which can be directly used to train the student model for the entire market (which is celled here the \textit{overall student model}). 

The new training data can also be further divided into disjoint subsets by applying a market segmentation approach as discussed earlier. Consider a categorical feature $\boldsymbol{X}_s$ and its set of possible values, i.e., $\boldsymbol{S} = \{S_1, ..., S_Q\}$. The market segmentation approach based on feature $\boldsymbol{X}_s$ divides the entire new training set $D$ into $Q$ disjoint subsets, i.e., $D_{S_1}, D_{S_2}, ..., D_{S_Q}$, where $D_{S_q}$ represents a subset with each $X_s = S_q, q = 1, ..., Q$. Consider now another categorical feature $\boldsymbol{X}_v, v \neq s$ with possible values $\boldsymbol{V} = \{V_1, ..., V_L\}$. Then, the market segmentation approach based on $\boldsymbol{X}_s$ and $\boldsymbol{X}_v$ divides the new training set $D$ into $Q \times L$ disjoint subsets, and a subset $D_{S_q, V_l}$ indicates a subset with $X_s = S_q$ and $X_v = V_l$. These definitions can be naturally extended to market segmentation based on multiple features and to non-categorical features by partitioning its domain. 

Observe that, when a decision tree is used as a student model to train on the whole data set, a similar phenomenon may be observed as its splits are essentially partition the data into disjoint subsets. However, manual market segmentation offers several advantages. First, decision trees are restricted to binary splits, but it may be more desirable to segment the domain into more subsections depending on the specific feature used. Moreover, market segmentation is usually conducted using some expert knowledge as to which feature to use, and the resulting subsets ideally exhibit more homogeneity, thus improving prediction accuracy and interpretation.

These subsets are then used to train different student models for different subpopulations (e.g., student model 1 for subset 1). Note that the model class of the students is typically determined by the modelers based on their preferences and needs. The evaluation of the the student models' performance, computes and compares two metrics, including $R^2$ (for fidelity) and out-of-sample prediction error (for accuracy). The resulting student models can be used for drawing inferences, extracting knowledge, and facilitating decision making.

\section{Case Study}

The proposed methodological framework aims at guiding those using machine learning for modeling travel behavior in how to apply model distillation to achieve various explanations and insights. This framework was implemented in the following case study of Ann Arbor, Michigan to illustrate the process and show the feasibility of the approach.

\subsection{The Data and Pre-Processing}

The data was collected from a stated-preference survey conducted at the University of Michigan, Ann Arbor, in 2017. It collected responses from the participants (including faculty, staff, and students) about the estimated travel time, travel cost, and wait time for their commuting trip from one of following travel modes: \textit{Car}, \textit{Walk}, \textit{Bike}, and \textit{Bus}. Then, the survey presented the participants a new mobility option (i.e., \textit{MOD Transit}) that would replace the existing bus system, and asked them to re-evaluate their mode choices in various state-choice experiments with different levels of service for MOD Transit. The detailed description for the survey design is included in \citep{YAN2018}. 

This paper aims at evaluating the factors underlying individuals' intention to switch to the new travel mode, i.e., MOD Transit. In other words, the goal is to model modal shift from people's current travel mode to the new MOD Transit: Other travel behavior changes, such as switching from Walk to Bike, is not of interest here. The response variable is binary with 1 indicating the decision of {\em switching to MOD Transit} and 0 denoting the decision of {\em not switching to MOD Transit}. As MOD Transit 
is deemed as a new travel mode, hence, individuals who currently use bus services are considered as switching to MOD Transit if she chooses the new travel mode.

\begin{table}[!t]
\centering
\caption{Statistics for features and response variable.}
\setlength{\tabcolsep}{7pt}
\resizebox{1\textwidth}{!}{
     \begin{tabular}{ l | l | l l l l l l }
       \hline
       \textbf{Variable} & \textbf{Description} & \textbf{Category} & \textbf{\%} & \textbf{Min} & \textbf{Max} & \textbf{Mean} & \textbf{SD}  \\[0.5ex]  \hline
       \textit{Response Variable} &&&&&&& \\[0.5ex]
       Switching Choice  & & MOD Transit (denoted by 1)  & 35.28 &        &       &  & \\
       &   & Not MOD Transit (denoted by 0) & 64.72  &        &       &  & \\
       & &&&&&& \\
       \textit{Input Features} &&&&&&& \\[0.5ex]
       TT\_Drive & Travel time of driving (min) &  &  & 2.00 & 40.00 & 15.21 & 6.62
       \\
       TT\_Walk & Travel time of walking (min)  &  &  & 3.00 & 120.00 & 32.30 & 23.08
       \\
       TT\_Bike & Travel time of biking (min) &  &  & 1.00 & 55.00 & 15.34 & 10.45
       \\
       TT\_MOD & Travel time of using MOD transit (min)    &  &  & 6.20 & 34.00 & 18.68 & 4.75
       \\
       Wait\_Time & Wait time for MOD (min) &  &  & 3.00 & 8.00  & 5.00 & 2.07
       \\
       Transfer & Number of transfers in MOD &  &  & 0.00 & 2.00  & 0.33 & 0.65
       \\
       Rideshare & Number of additional pickups in MOD &  &  & 0.00 & 2.00  & 1.11 & 0.82
       \\
       Income& Income level
       &  &  & 1.00 & 6.00  & 1.93 & 1.34 \\
       Bike\_Walkability& Importance of bike- and walk-ability   &  &  & 1.00 & 4.00  & 3.22 & 0.95 \\
       MOD\_Access& Ease of access to MOD  &  &  & 1.00 & 4.00  & 3.09 & 1.02  \\
       CarPerCap & Car per capita  &  &  & 0.00 & 3.00  & 0.53 & 0.48
       \\
       Female & Female or Male  & Female (denoted by 1)   & 56.32   & & & & \\
       & & Male (denoted by 0) & 43.68  & & & & \\
       Student & Students or faculty/staff  & Student (denoted by 1) & 73.52 & & & & \\
       && Faculty or staff (denoted by 0) & 26.48  & & & & \\
       Current\_Mode  & Current travel mode & Car (denoted by 5) & 16.68 & & & & \\
       && Walk (denoted by 6) & 40.41  & & & & \\
       && Bike (denoted by 7) & 8.25 & & & & \\
       && Bus (denoted by 8) & 34.65  & & & & \\
       [1ex]
       \hline
    \end{tabular}
    }
     \label{tab:var}
\end{table}

In total, 8,141 observations were collected from 1,163 individuals, as each individual conducted seven state-choice experiments. The statistics for 14 selected features and the response variables are provided in Table \ref{tab:var}. Note that the \textit{Income} feature indicates the annual household income for faculty/staff and the annual living expenditure for students, and the detailed definition of the six different income levels for faculty/staff and students is presented in Table \ref{tab:INC}.

\begin{table}[ht!]
\centering
\caption{Definition of income levels for faculty, staff, and students.}
\footnotesize
\begin{tabular}{ccc}
\hline
Income & Annual household income of faculty/staff & Annual living expenditure of students \\ \hline
1      & less than \$50,000                       & less than \$20,000                    \\
2      & \$50,000 -- \$74,999                       & \$20,000 -- \$34,999                    \\
3      & \$75,000 -- \$99,999                       & \$35,000 -- \$49,999                    \\
4      & \$100,000 -- \$149,999                     & \$50,000 -- \$74,999                    \\
5      & \$150,000 -- \$199,999                     & \$75,000 -- \$99,999                    \\
6      & \$200,000 or more                        & \$100,000 or more                     \\ \hline
\end{tabular}
\label{tab:INC}
\end{table}

Before training the teacher and student models, the data was checked for multicollinearity: All features have a variance inflation factor of no more than five (a commonly-used threshold), indicating that multicollinearity is not an issue in this study. As each individual filled out seven experiments, the test set consisted of a randomly sampled  experiment for each individual (i.e., 1,163 instances are held out for testing the out-of-sample prediction capability). The remaining 85.7\% of the data (with 6,978 instances) is used for training. As discussed earlier, additional synthetic data should be generated to enhance the stability of decision trees; however, our original data is relatively high-dimensional and the kernel density method (used in \citep{zhou2018approximation,gibbons2013cad}) is difficult to be applied here to generate additional data. Additionally, the size of our dataset seems to be adequate to fit relatively stable decision trees. Therefore, instead of generating synthetic data, we randomly hold out 30\% of the training data as the ``additional'' data and use the remaining 70\% to fit the teacher model (i.e., 60\% of the entire dataset). This makes it possible to test out the proposed methodological framework (see Figure \ref{fig:framework}) and evaluate its performance in a holistic manner. The process is repeated for 30 times to generate mean estimate and its standard deviations as shown in Table \ref{tab:Comp}. 

\subsection{Teacher Model: Boosting Trees Model}

When training a teacher model, one is advised to select a set of candidate machine-learning models first and then compare the model performance through cross validation or holdout testing. \citet{zhao2019modeling} used the same dataset\footnote{A slight difference is that, in this paper, three binary features used in \citep{zhao2019modeling}, including Current\_Mode\_Car, Current\_Mode\_Walk, and Current\_Mode\_Bike, are combined into a four-level feature, i.e., Current\_Mode.} and identified the boosting trees model as the best-performing model among seven different machine learning classifiers, including logistic regression, Naive Bayes, classification and regression tree (CART), bagging trees, boosting trees, random forests, and artificial neural networks. Therefore, in this paper, the boosting trees model is used as the teacher model. 

Boosting trees is one of the most popular tree-based ensemble methods that aims at forming robust, stable, and accurate classifiers compared to a single decision tree \citep{friedman2001greedy, friedman2001elements}. For a classification problem, the boosting trees model generates a sequence of decision trees, where each successive decision tree seeks to improve the classification accuracy of the previous one. For the boosting trees model, the final predicted outcomes are based on a weighted voting among all the trees. This paper applies the gradient boosting method to create the boosting trees model \citep{friedman2001greedy}. Specifically, the hyperparameters are selected as: 500 trees are used, with the shrinkage parameter set to 0.062 and the interaction depth to 45, and the minimum number of observations in leaves is 10. The R package \textit{gbm} \citep{gbm} is used for conducting the modeling and analysis. 

Then, according to the methodological framework (see Figure \ref{fig:framework}), the class labels of the training data are discarded, and the teacher model is used to make soft predictions (i.e., the switching probabilities) for the same training data. The input features together with the predicted outcome of the teacher model form the new training data, which will be used to train student models.

\subsection{Student Models}

As mentioned, this paper focuses on decision trees to fit student models. Decision trees recursively partition the feature space into sub-regions until some stopping rule is applied \citep[p.~305]{friedman2001elements}. Decision trees can be used for both regression and classification, and in this paper, the CART algorithm is applied through the R package \textit{tree} \citep{tree}. The decision tree is good at modeling the nonlinearity between the outcome and features as well as capturing the interactions between input features. Moreover, decision trees can present a straightforward visualization and offer good interpretability (as long as the tree size is not too large). Therefore, to control the complexity of the decision trees for transparent interpretations, we limit the tree size (i.e. the number of terminal nodes or leaves) to be no more than 10. However, as discussed in \citet{molnar}, the decision tree lacks smoothness in predictions or stability of the tree structure; in addition, the trees cannot model linear relationships efficiently since the linear relationships are approximated by splits (i.e., a step function).

In this paper, multiple decision-tree-based student models are trained for the entire market and different subgroups (categorized by the current travel mode or by the income levels). As a comparison, we fit a \textit{benchmark decision tree model} using the same data (i.e., 60\% of the entire dataset) that is used to train the teacher model.

\subsection{Model Comparison}

This section compares the performance of the student models, the benchmark decision tree model, and the teacher model, in terms of fidelity and accuracy. Additionally, it investigates the model interpretation of the overall student model and the benchmark decision tree model, and further looks into the explanations of the student models for different subpopulations. 

Some of the key observations are summarized here: 1) The overall student model can improve performance compared to the benchmark decision tree; 2) Segmented by income levels, the resulting student models can be significantly improved compared to the overall student model and the benchmark decision tree; 3) The interpretation gained from the overall student model is similar but has differences (more sophisticated) than that obtained from the benchmark decision tree; and 4) The student models for different income levels present various decision rules, leading to rich insights for policy design.

\subsubsection{Fidelity and Accuracy}

Table \ref{tab:Comp} shows the comparison of student models, the benchmark decision tree, and the teacher model, with $R^2$ quantifying how the student model is faithful to the teacher's predictions and the out-of-sample accuracy capturing the predictive capability of different models. 

\begin{table}[ht!]
\caption{Comparison of student models, benchmark decision tree, and teacher model: Higher $R^2$ and higher out-of-sample accuracy (denoted by ACC) are better. Each experiment is repeated for 30 times.}
\centering
\small
\begin{tabular}{rllllllll}
\hline
\multicolumn{1}{l}{\multirow{2}{*}{Scenario}} & \multicolumn{2}{l}{$R^2$ of Student} & \multicolumn{2}{l}{ACC of Student} & \multicolumn{2}{l}{ACC of Benchmark DT} & \multicolumn{2}{l}{ACC of Teacher} \\
\multicolumn{1}{l}{}                          & Mean                 & SD                  & Mean                  & SD                    & Mean                  & SD                   & Mean                  & SD                    \\ \hline
\multicolumn{1}{l}{Entire population}         & 0.3018               & 0.0110              & 0.7333                & 0.0126                & 0.7197                & 0.0113               & 0.8661                & 0.0105                \\ [1ex]
Car                             & 0.2445               & 0.0228              & 0.6582                & 0.0353                & 0.5892                & 0.0333               & 0.8326                & 0.0266                \\
Walk                            & 0.3100               & 0.0198              & 0.8130                & 0.0172                & 0.7962                & 0.0145               & 0.8696                & 0.0142                \\
Bike                            & 0.4941               & 0.0576              & 0.8969                & 0.0260                & 0.8733                & 0.0252               & 0.9205                & 0.0297                \\
MOD Transit                     & 0.2954               & 0.0199              & 0.6991                & 0.0124                & 0.6566                & 0.0228               & 0.8652                & 0.0168                \\ [1ex]
Income = 1                                    & 0.3594               & 0.0152              & 0.7516                & 0.0168                & 0.7383                & 0.0116               & 0.8705                & 0.0109                \\
Income = 2                                    & 0.3383               & 0.0166              & 0.7479                & 0.0185                & 0.7074                & 0.0226               & 0.8549                & 0.0228                \\
Income = 3                                    & 0.5053               & 0.0361              & 0.7492                & 0.0439                & 0.6742                & 0.0301               & 0.8587                & 0.0352                \\
Income = 4                                    & 0.5647               & 0.0288              & 0.7937                & 0.0384                & 0.6706                & 0.0337               & 0.8675                & 0.0318                \\
Income = 5                                    & 0.7958               & 0.0382              & 0.8819                & 0.0461                & 0.7229                & 0.0438               & 0.8971                & 0.0490                \\
Income = 6                                    & 0.6156               & 0.0432              & 0.8029                & 0.0474                & 0.7138                & 0.0428               & 0.8645                & 0.0422                \\ \hline
\end{tabular}
\label{tab:Comp}
\end{table}

The results of $R^2$ are shown in the first column of Table \ref{tab:Comp}. Some of the student models have relatively high $R^2$ statistics, i.e., well above 0.5; however, some of them have $R^2$ less than 0.3, showing relatively poor variance explanation for the teacher. However, it remains unclear how to determine the best threshold for $R^2$ in order to be confident about the student model's interpretation \citep{molnar}. An important observation is that the overall student model has a fair $R^2$ value (0.3018), but by splitting the entire population by income levels and fit different student models, the corresponding $R^2$ values are all improved, especially for income levels 3 through 6. In contrast, when disaggregating the population by the current travel mode choice, the corresponding $R^2$ has mixed outcomes: It is improved for Walk and Bike, but becomes worse for Car and MOD Transit. One possible explanation is that people who are currently using motorized modes are presenting very heterogeneous switching behavior and cannot be fully captured by a single decision tree; similar results can be found in \citet{zhao2019modeling}. 

For the out-of-sample accuracy, the overall student model (0.7333) is showing significantly better performance than the benchmark decision tree (0.7197), according to the $p$-value (2.152 $\times 10^{-5}$) of a one-sided $t$-test of the mean estimates for the benchmark decision tree and the overall student model. That is to say, by applying model distillation, the student decision tree can make better predictions than the benchmark decision tree. This is a reasonable phenomenon to expect as the teacher model is able to capture more complex signal in the data and by mimicking its predictions, the student decision tree can extract additional  information from an enriched training data and more meaningful target values \citep{hinton2015distilling, gibbons2013cad}. Furthermore, by segmenting the entire population by the current travel mode and the income level, the resulting student models all show higher out-of-sample accuracy than the benchmark decision tree. The main reason is that the benchmark decision tree is fitted for the entire population while these student models only focus on specific market segments and offer more specialized interpretations. Notably, some of the student models (e.g., Bike and Income = 5) can achieve similar predictive accuracy compared to the teacher model and produce high $R^2$ values (showing its high faithfulness to the teacher's predictions), so they may be used to replace the teacher model in these cases.

Another finding is that $R^2$ and the out-of-sample predictive accuracy are not necessarily positively correlated: Even though $R^2$ for Income = 3 is significantly higher than that for Income = 1, the corresponding predictive accuracy for Income = 3 is lower. This is probably due to overfitting. Furthermore, $R^2$ for Walk is not high (0.3100) and it is high (0.5647) for Income = 4, but Walk and Income = 4 have the same level of out-of-sample predictive accuracy (0.8130 and 0.7937, respectively).


\subsubsection{Model Interpretation}

To interpret the student models and compare their interpretation with that gained from the benchmark decision tree model, we present the visualization of the fitted decision trees from a random realization.

\begin{figure}[!t]
    \centering
    \begin{subfigure}[b]{0.78\textwidth}
        \centering
        \includegraphics[height=2.0in]{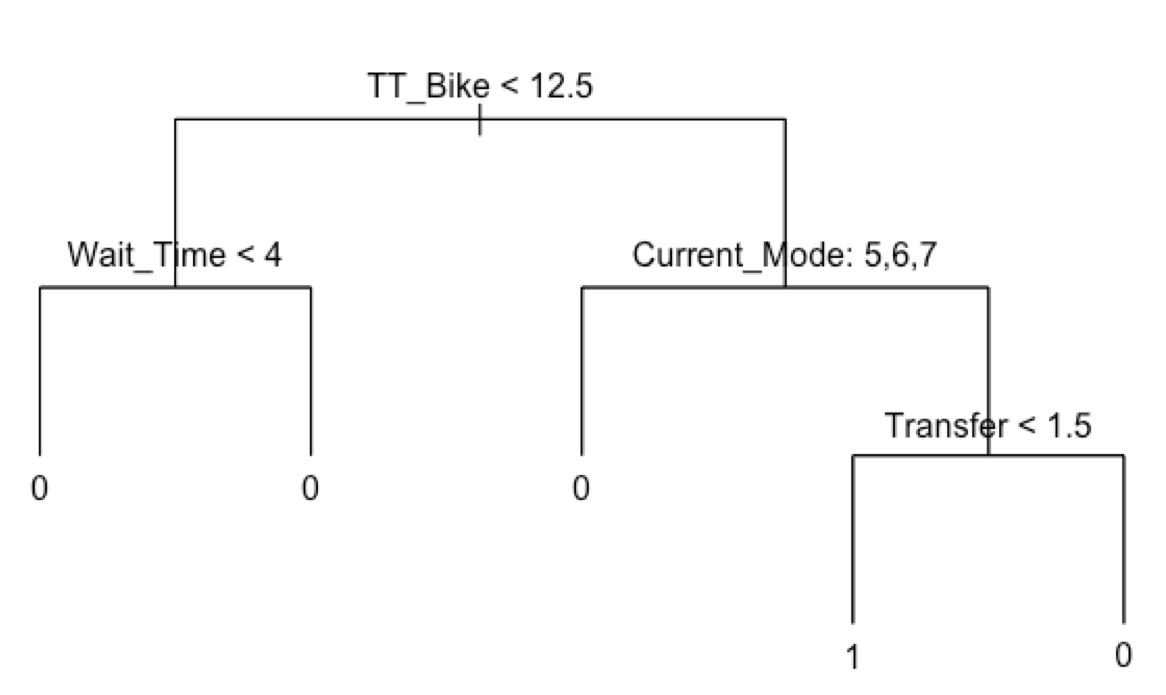}
        \caption{Benchmark decision tree: The leaves' values are classes (switching: 1; not switching: 0)}
        \label{fig:benchmark}
    \end{subfigure}%
    
    \begin{subfigure}[b]{0.78\textwidth}
        \centering
        \includegraphics[height=2.0in]{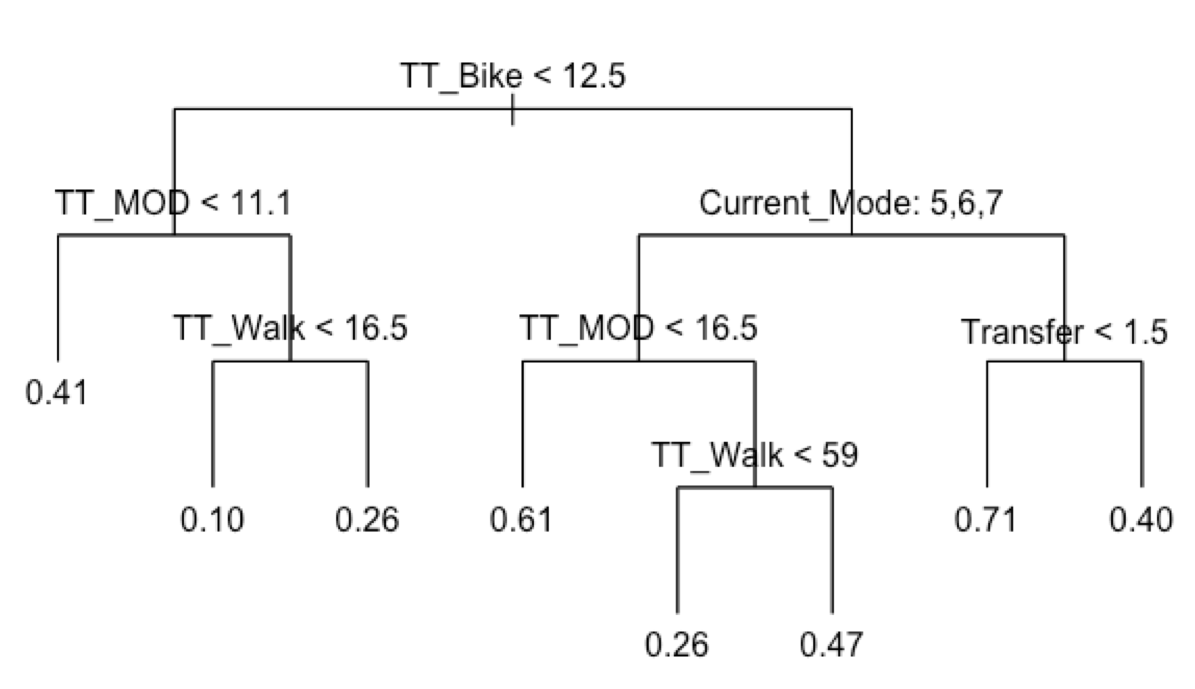}
        \caption{Overall student model: The leaves' values are switching probabilities (if less than 0.50, indicating not switching)}
        \label{fig:mimic}
    \end{subfigure}%
    \caption{Visualization for benchmark decision tree and overall student model.}
    \label{fig:DT_all}
\end{figure}

Figure \ref{fig:DT_all} presents the benchmark decision tree and the overall student model. Note that the benchmark decision tree is trained on the original training data with the response variable being binary, so the fitted tree is a classification tree; on the other hand, the overall student model is trained on the new training data with the response variable being the switching probability predictions provided by the teacher model, so the fitted tree is a regression tree. Thus, according to the benchmark decision tree (see Figure \ref{fig:benchmark}), we can find that only when the biking time is more than 12.5 min, the current travel mode is Bus, and the number of transfers is less than 1.5, the travelers will switch to MOD Transit. Figure \ref{fig:mimic} shows the overall student model for the entire population. If we transfer the switching probability to a binary outcome (i.e., if the switching probability is less than 0.50, it indicates not switching, and vice versa), then we find that under two sets of decision rules people will switch to MOD Transit: One is the same set of rules as the benchmark decision tree; the other is when the biking time is over 12.5 min, the current mode is driving, walking, or biking, and the travel time of MOD Transit is less than 16.5 min. Apparently, the switching logic presented by the benchmark decision tree is too simple, indicating no people will switch to MOD Transit other than the existing bus users. In contrast, the overall student model presents two different sets of decision rules, and one of them shows how non-transit users can be attracted to the new travel mode, i.e., MOD Transit, and this insight is particularly of interest for transportation planners. To summarize, the interpretation obtained from the overall student tree shares many similarities with the benchmark decision tree, but offers new and more sophisticated insights compared to that gained from its counterpart.

\begin{figure}[!t]
    \centering
    \begin{subfigure}[b]{0.78\textwidth}
        \centering
        \includegraphics[height=2.0in]{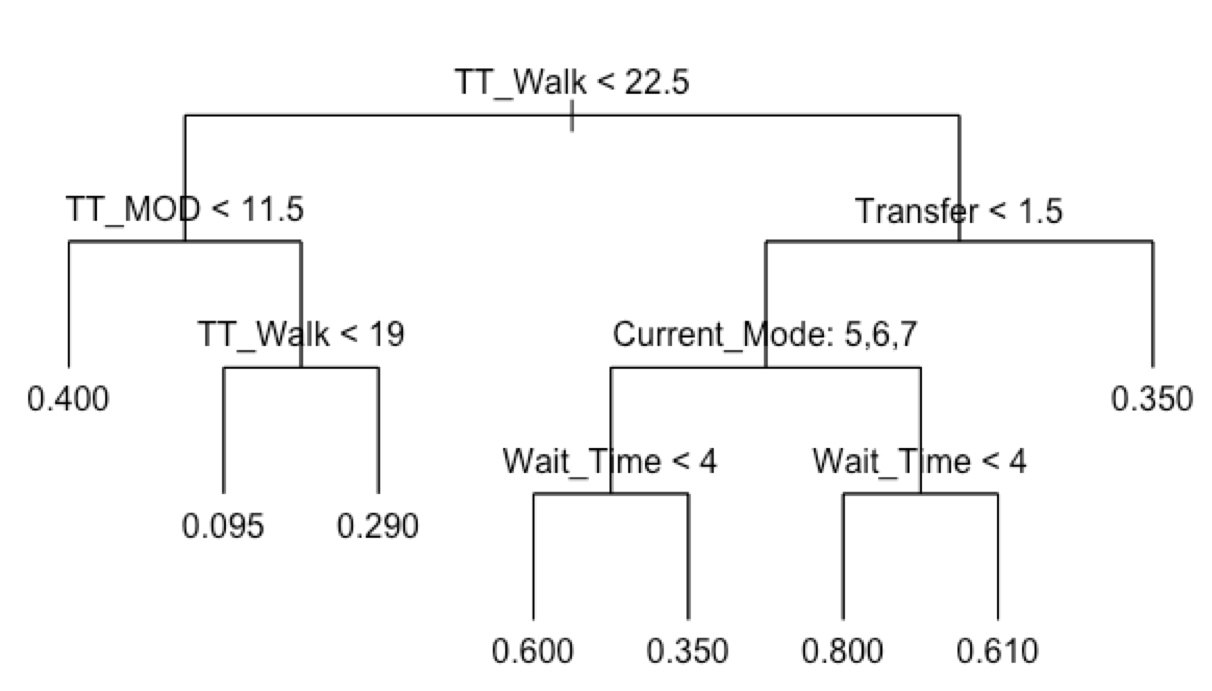}
        \caption{Student model for income = 1}
        \label{fig:inc1}
    \end{subfigure}%
    
    \begin{subfigure}[b]{0.78\textwidth}
        \centering
        \includegraphics[height=2.0in]{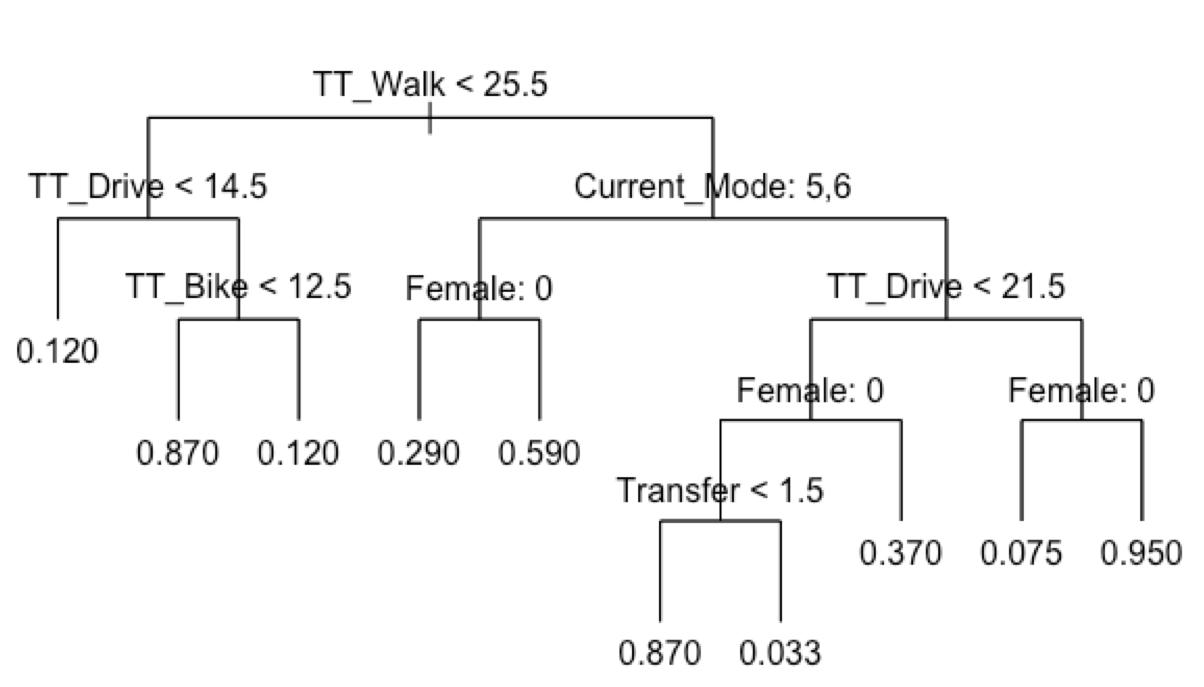}
        \caption{Student model for income = 3}
        \label{fig:inc3}
    \end{subfigure}%

    \begin{subfigure}[b]{0.78\textwidth}
        \centering
        \includegraphics[height=2.0in]{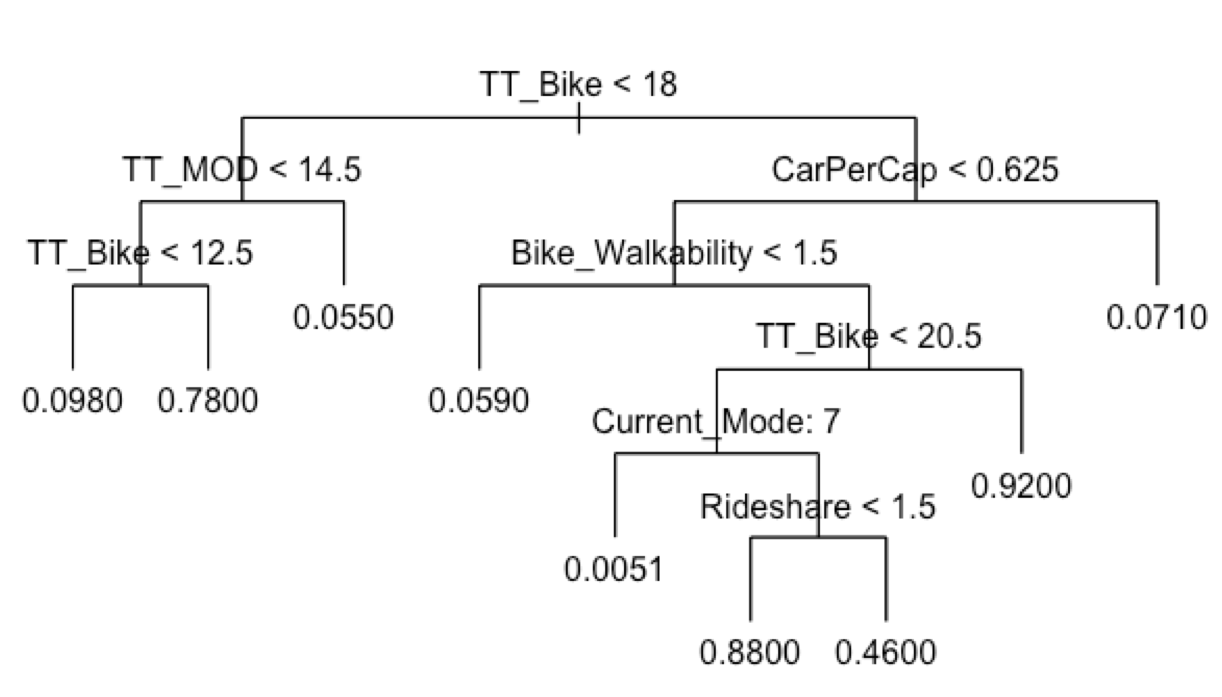}
        \caption{Student model for income = 5}
        \label{fig:inc5}
    \end{subfigure}%
    \caption{Visualization for student models for different income levels.}
    \label{fig:income}
\end{figure}

Figure \ref{fig:income} illustrates the student models for income = 1, 3, 5. These three models share some similarities: For example, features such as travel times for different modes and the current travel mode are used in all the three decision trees to determine splits. Despite their similarities, they have much more differences. For the student model for income = 1 (see Figure \ref{fig:inc1}), we can find that the decision rules for switching to MOD Transit are very similar to those for the overall student model with some variations; in particular, TT\_Walk instead of TT\_Bike is used at the top node for splitting, and the waiting time plays an important role in the decision making process for the low-income population. The student model for income = 3 resembles the student model for income = 1, but becomes more complex and has Female as an important feature for splitting. On the other hand, the student model for income = 5 becomes rather different from the student models for income = 1, 3; for the high-income population, in addition to travel times and current mode, people consider car ownership, the importance of bike- and walk-ability, and the number of additional pickups in MOD to decide if MOD Transit is a preferred mode for commuting. From these observations above, we may conclude that people with different income levels have very diverse decision-making processes for switching or not to MOD Transit. These insights can help design better-targeted policies to promote the modal shift from private vehicles to MOD Transit.

\section{Discussion and Conclusion}

This paper applied and extended model distillation to interpret black-box travel mode choice models. By compressing the knowledge from complex teacher models into understandable and interpretable student models, model distillation and its proposed extension can help transportation planners and engineers gain valuable insights for decision-making. In particular, the integration of model distillation with market segmentation generates unique insights that can be translated into more focused policies. The student models have a tree structure with clear visualizations and transparent interpretations. For agencies that plan to use black-box machine-learning methods to model travel behavior, model distillation can serve as a decision-support tool for policy makers to extract the underlying prediction mechanism of the black-box model, understand the relationships between the input features and outcomes, conduct policy analysis to assess different strategies, and plan interventions accordingly. 

From an accuracy standpoint, the results show that student models can achieve better performance than the benchmark decision tree models. By using the predicted soft target (instead of using the binary hard label) to train the student models, they can learn more information about the prediction mechanism of the teacher model and thus achieve better out-of-sample predictive capability. To further support this argument, we fitted another benchmark decision tree model using the same data (i.e., 85.7\% of the entire dataset) that is used to train the overall student model, but with the observed switching choice as the response variable. The results show that the out-of-sample accuracy of the new benchmark decision tree model is 0.7226 (versus 0.7333 for the overall student model). A one-sided $t$-test was also conducted and the results show that the overall student model is significantly better than the benchmark decision tree model ($p$-value = $1.821 \times 10^{-4}$). This shows that model distillation can help student models directly learn from the teacher model (via the predicted soft target) and thus achieve better performance. In addition, with the teacher model, additional synthetic data can be generated to train student models, which can not only help further improve the predictive capability of student models but also make the student decision trees more stable \citep{zhou2018approximation}. 

More accurate student models, especially student models for subpopulations, may be used to replace the black-box teacher model for undertaking the forecasting task as well as offering crisp explanations. For example, the learned student models can be used as an input into some activity-based travel demand models, such as \citep{arentze2004learning,pendyala1998application}, to improve their predictive accuracy.



\section{Acknowledgements}

This research was partly funded by the Georgia Institute of
Technology, the Michigan Institute of Data Science (MIDAS), and Grant
7F-30154 from the Department of Energy.

\bibliographystyle{trb}
\bibliography{trb.bbl}

\end{document}